\documentclass{article} 
\usepackage{times}
\usepackage[round]{natbib}

\usepackage{hyperref}
\usepackage{url}
\usepackage[margin=1in]{geometry}
\usepackage[margin=1in]{geometry}

\usepackage{caption}
\usepackage{multirow}
\usepackage{amsmath}
\usepackage{subfiles}
\usepackage{array}
\usepackage{graphicx}
\usepackage{subfigure}
\usepackage{amsthm}
\usepackage{wrapfig}
\usepackage{booktabs} 
\usepackage{subcaption}
\usepackage{makecell}
\usepackage{adjustbox}
\usepackage{wrapfig}
\usepackage{multirow} 

\title{MemArchitect: A Policy Driven Memory Governance Layer}
\date{}

\author{%
Lingavasan Suresh Kumar\thanks{Email: \texttt{lsuresh4@asu.edu}} \\
Arizona State University
\and
Yang Ba\thanks{Email: \texttt{Yangba@asu.edu}} \\
Arizona State University
\and
Rong Pan\thanks{Email: \texttt{Rong.Pan@asu.edu}} \\
Arizona State University
}

\begin{document}


\maketitle

\begin{abstract}

As Large Language Models (LLMs) transition into persistent agents, they face
a critical “\textbf{Governance Gap.}” Standard Retrieval Augmented Generation (RAG)
treats memory as a passive storage bucket, lacking the controls necessary to
manage contradictions, enforce privacy, or prevent “zombie memories” (outdated
facts) from polluting the context window. We introduce \textbf{MemArchitect}, a
distinct governance layer that decouples memory lifecycle management from model
weights. By enforcing explicit, rule based policies ranging from memory decay to
user privacy enforcement, \textbf{MemArchitect} addresses these challenges. We demonstrate
that governed memory outperforms raw memory, highlighting the necessity of
governance in autonomous agentic environments.

\end{abstract}

\section{Introduction}

As Large Language Models (LLMs) evolve from stateless chatbots to autonomous agents, they face a critical failure mode: \textit{Memory Hallucination} \citep{Iyidogan2025}. Unlike standard hallucinations, these occur when an agent retrieves conflicting or outdated facts from its own history (e.g., retrieving ``User is single'' from 2024 alongside ``User is married'' from 2025). Simply expanding context windows exacerbates \emph{context pollution}, rendering agents unreliable for long-horizon tasks with frequent state changes.

The agentic AI field currently addresses these challenges through three distinct architectural paradigms: operating systems, efficient streaming, and structured retention. First, frameworks such as MemGPT \citep{Packer2024} and MemOS \citep{Li2025} treat the LLM as an operating system, using virtual memory paging to manage context, while recent innovations such as Titans \citep{Behrouz2024} propose ``neural memory'' modules that allow models to learn context at test time. Second, to sustain these lifelong interactions, approaches such as SimpleMem \citep{simplemem2025} and LLMLingua \citep{Jiang2023} focus on semantic compression, building on the foundational Attention Sink \citep{xiao2023streamingllm} mechanisms that enable stable processing of infinite token streams. Finally, the focus on semantic organization is evolving rapidly; while Generative Agents \citep{Park2023} and MemoryBank \citep{Zhong2023} introduced reflection and decay to mimic human retention, newer approaches such as A-MEM \citep{Xu2025} and the Random Tree Model of Meaningful Memory \citep{Zhong2025} advocate using the hierarchical Zettelkasten-style evolutionary structures to replace flat logs. Commercial-grade implementations, such as Mem0 \citep{Chhikara2025}, further extend this approach by employing graph-enhanced retrieval to persist user preferences across sessions.

However, these systems suffer from a distinct ``Governance Gap.'' They treat memory as a passive, append-only log, lacking mechanisms to resolve state contradictions, enforce ``Right-to-be-Forgotten'' compliance, or actively prune obsolete data based on utility.
Therefore, we propose MemArchitect, a governance middleware that shifts the paradigm from passive memory retrieval to active adjudication. Unlike standard RAG systems that treat memory as a static storage bucket, MemArchitect interposes a “Triage \& Bid” economy between the user and the agent, where every piece of information must actively compete for the context window. This policy engine enforces a unified governance protocol in four critical domains: \emph{Lifecycle \& Hygiene, Consistency \& Truth, Provenance \& Trust, and Efficiency \& Safety,} ensuring that the agent’s context is not only retrieved but proactively managed for decay, factuality, lineage, and compliance standards.

\section{The Governance Engine: Policy Layer}
\label{sec:methodology}

\subsection{The Policy Suite}
To transition memory from a passive artifact to an active cognitive resource, we establish a unified governance protocol anchored in four foundational pillars: \textbf{Lifecycle}, \textbf{Consistency}, \textbf{Retrieval}, and \textbf{Efficiency}. We argue that for an autonomous agent to function indefinitely without succumbing to ``context collapse,'' it must systematically address the entropy of time, the drift of truth, the complexity of access, and the scarcity of computing resource. By governing these four dimensions, we ensure that the memory store operates not merely as a database, but as a highly alerted environment where information actively competes for relevance. ({For the complete mathematical formulations and implementation details of all  policies, please see Appendix A.})
\begin{enumerate}
    \item \textbf{Lifecycle \& Hygiene (The ``Forgetting'' Engine):}
    \begin{itemize}
        \item \textit{FSRS Decay:} We replace exponential decay with the \textit{Free Spaced Repetition Scheduler} (FSRS v4)\cite{ye2022fsrs}. Memory decay mimics a biological forgetting curve where retrievability ($R$) depends on stability ($S$) and time ($t$).
        \item \textit{Entropy-Triggered Consolidation:} A background process monitors information density. High redundancy (Entropy Ratio $< 0.4$) triggers a cleanup cycle where fading episodic memories ($0.3 \le R \le 0.7$) are compressed into stable semantic facts, and ``forgettable'' noise ($R < 0.3$) is actively pruned.
    \end{itemize}

    \item \textbf{Consistency \& Truth (The ``Utility'' Engine):}
    \begin{itemize}
        \item \textit{The Reflection Loop (Kalman Filter):} We treat the retrieval utility as an alert signal. A Kalman Filter tracks the ``Trust Score'' ($U$) of every memory. Useful memories gain inertia ($U \uparrow$), while hallucinations are mathematically down-weighted.
        \item \textit{Relevance Discriminator (Veto Gate):} A secondary Cross-Encoder acts as a ``Veto'' gate. Even if a memory passes the vector search, it is discarded if it does not meet a strict entailment threshold (i.e., $< 0.1$). This prevents highly similar but irrelevant noise from polluting the context.
    \end{itemize}

    \item \textbf{Adaptive Retrieval (The ``Auction'' Engine):}
    \begin{itemize}
        \item \textit{Multi-Hop Decomposition:} Complex queries are broken into sub-questions (e.g., ``Where is Tokyo?'' + ``Capital of that country?'') to retrieve disjointed facts.
        \item \textit{Adaptive Scoring:} Queries are classified (Fact, Temporal, Reasoning) for dynamically tuning retrieval weights, where old memories are penalized for immediate reasoning tasks ($\lambda=1.0$), and facts are treated as timeless ($\lambda=0$).
        \item \textit{Hebbian Graph Expansion:} To bridge non-semantic gaps, we implement a Hebbian Graph Expansion policy, inspired by the foundational principle of associative learning \citep{hebb1949organization}, where ``memories that fire together are wired together." That is, if Memory A and B are frequently used together (e.g., $P(B|A) > 0.7$), retrieving A automatically pulls B into context.
    \end{itemize}

    \item \textbf{Efficiency (The ``Budget'' Engine):}
    \begin{itemize}
        \item \textit{Adaptive Token Budgeting:} The system dynamically allocates the context window. High-confidence retrieval sets a ``Reasoning Reserve''(30\%) for the LLM to think, while low-confidence retrieval maximizes the ``Recall Reserve'' (10\%) to seek for facts.
    \end{itemize}
\end{enumerate}

\subsection{The Governance Cycle \& Dependencies}
\label{subsec:gov_cycle}

\begin{figure}[ht]
    \centering
    \includegraphics[width=1.0\linewidth]{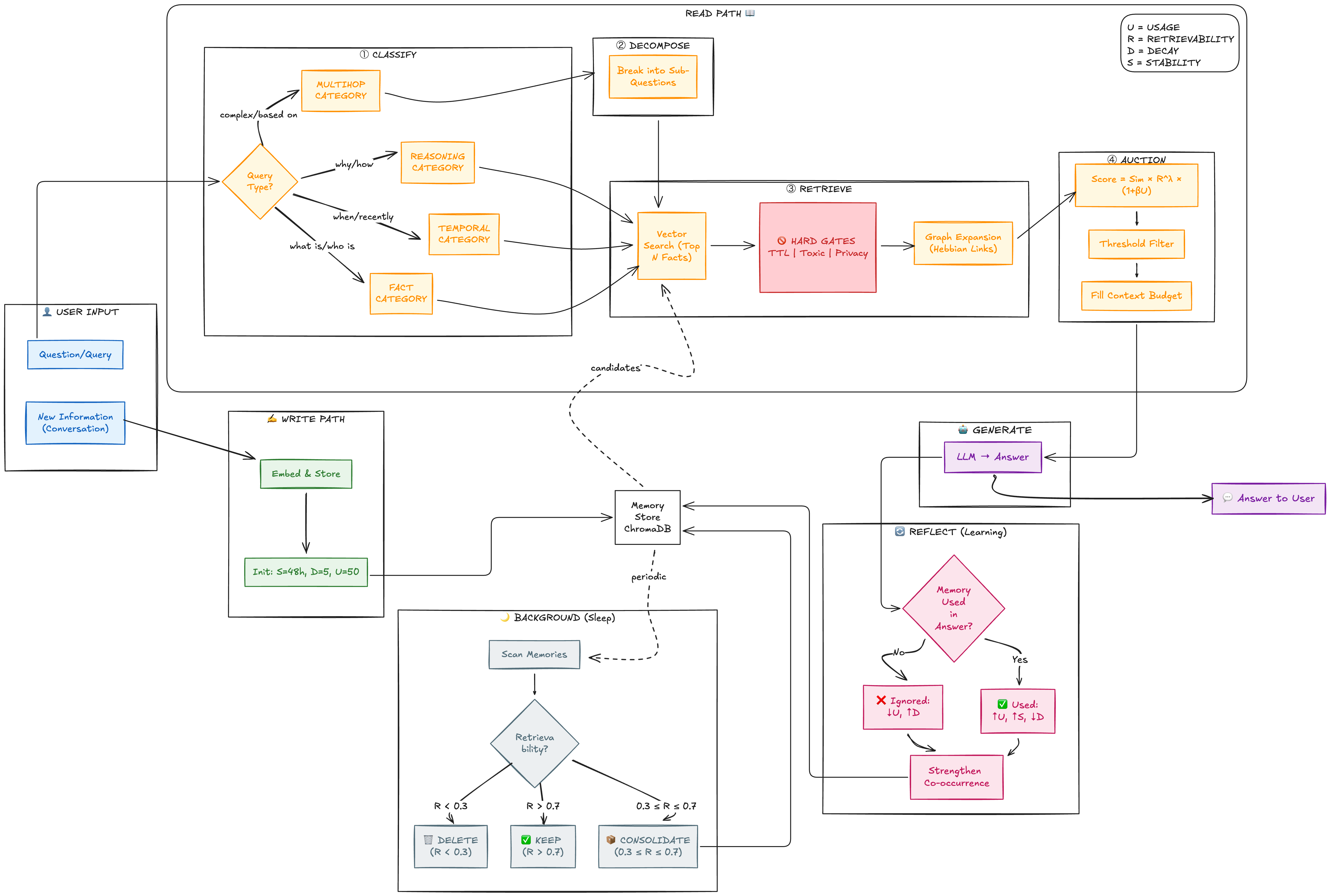}
    \caption {\textbf{The MemArchitect Governance Workflow.} The system operates as a continuous closed loop across three paths:
    \textbf{(1) Read Path (Top):} Incoming queries are classified by intent to customize the search. Memories are retrieved, filtered for safety, and ranked by value to fit the limited context budget.
    \textbf{(2) Reflect Path (Right):} A feedback loop detects which memories were actually used to generate the answer, strengthening useful information while penalizing distractions.
    \textbf{(3) Background Path (Bottom):} During idle cycles, a maintenance process automatically cleans the memory store by deleting forgotten noise and compressing fading details into permanent facts.}
        \label{fig:workflow}
\end{figure}

These policies do not operate in isolation; they form a tightly coupled functional loop, as illustrated in Figure \ref{fig:workflow}. The architecture splits the agent's operation into three distinct execution paths:
\begin{enumerate}
    \item \textbf{Read Path (The Retrieval Auction):} Upon receiving a query, the \textit{Classifier} first determines the active scoring logic ($\lambda, \beta$) based on intent (e.g., Fact vs. Reasoning). Simultaneous \textit{Decomposition} breaks complex queries into atomic sub-questions. Candidates then enter the \textit{Auction}, where they are ranked by their computed scores, rigorously filtered by the \textit{Discriminator} (Cross-Encoder) for semantic entailment, and expanded via \textit{Hebbian Links} to retrieve associated context before filling the \textit{Adaptive Token Budget}.

    \item \textbf{Reflect Path (The Feedback Loop):} Post-generation, the system closes the learning loop. \textit{Usage Detection} algorithms analyze the generated output to identify which specific memories contributed to the answer. This binary feedback signal triggers immediate updates to the \textit{Kalman Utility} ($U$) state (reinforcing trust) and the \textit{FSRS Stability} ($S$) metric (to reset the forgetting curve).

    \item \textbf{Background Path (The Hygiene Engine):} During system idle time, the \textit{Consolidation} policy scans the vector store. Driven by the updated Stability scores, it enforces the lifecycle rules: low-retrievability noise is physically deleted, while fading episodic memories are compressed into stable semantic facts, ensuring the long-term health of the memory graph.
\end{enumerate}

\section{Experiment \& Analysis}

\subsection{Experiment Setup and Baseline}
MemOS (Memory Operating System) \citep{Li2025} and SimpleMem \citep{simplemem2025} capture two complementary extremes, respectively; i.e., system-level governance and algorithmic efficiency. They form natural baselines for evaluating recall quality and token efficiency on long-horizon benchmarks such as LoCoMo(\cite{maharana2024evaluatinglongtermconversationalmemory}). Qwen-3B and Llama-3.1-8B are used in our experiments and evaluated in accuracy. 

MemOS reframes LLM memory as a system-level resource (``MemCube") and manages its lifecycle through an OS-style module, called MemScheduler. It represents a maximalist, full-stack approach. Like MemOS, we treat memory as a governed resource. However, MemArchitect operates strictly as an external governance layer without modifying model parameters, making it lightweight and model-agnostic. SimpleMem targets token inefficiency using a three-stage pipeline -- compression, consolidation, and adaptive retrieval. It achieves strong efficiency by maximizing information density per token. In comparison, MemArchitect shares the efficiency goal but differs in mechanism. While SimpleMem relies on compression, we rely on \emph{policy-driven governance} including FSRS decay, utility tracking, and auctions.

\subsection{Quantitative Results}

\begin{table}[h]
\centering
\caption{Comparison of Memory Systems on LoCoMo-10 Benchmark}
\label{tab:merged_results}
\resizebox{\linewidth}{!}{%
\begin{tabular}{llcc}
\toprule
\textbf{Model} & \textbf{Task} & \textbf{Baseline (Acc)} & \textbf{MemArchitect (Acc)} \\
\midrule
\multirow{5}{*}{Qwen-3B} 
 & Single-Hop         & 79.4\% (SimpleMem) & 95.0\% \,(\,+15.6\%) \\
 & Multi-Hop          & 66.7\% (SimpleMem) & 95.8\% \,(\,+29.1\%) \\
 & Temporal Reasoning & 56.1\% (SimpleMem) & 95.3\% \,(\,+39.2\%) \\
 & Open Domain        & 67.1\% (SimpleMem) & 93.1\% \,(\,+26.0\%) \\
\midrule
\multirow{5}{*}{Llama-3.1-8B}
 & Single-Hop         & 74.0\% (MemOS)     & 53.2\% \,(\,-20.8\%) \\
 & Multi-Hop          & 36.0\% (MemOS)     & 40.1\% \,(\,+4.1\%) \\
 & Temporal Reasoning & 72.0\% (MemOS)     & 27.1\% \,(\,-44.9\%) \\
 & Open Domain        & 45.0\% (MemOS)     & 56.3\% \,(\,+11.3\%) \\
\midrule
 & \textbf{Average} & 62.04\% & \textbf{69.49\% (\,+7.45\%)} \\
\bottomrule
\end{tabular}%
}
\end{table}



Table \ref{tab:merged_results} demonstrates the superior performance of MemArchitect in most tasks and shows an aggregated improvement of \textbf{+7.45\%} accuracy across all tasks. When compared against the algorithmic baseline (SimpleMem), MemArchitect achieves a dominant sweep, confirming that policy-driven adjudication is superior to semantic compression for preserving context fidelity. SimpleMem's compression indiscriminately discards temporal markers, leading to poor performance in Temporal Reasoning, whereas our FSRS policy actively prioritizes them (95.3\% accuracy, a \textbf{+39.2\%} gain).

However, the lower half of the Table reveals a critical trade-off against the MemOS, which treats memory as an infinite OS resource, dominating in raw recall tasks (Single-Hop and Temporal) because it retains all logs. Our active decay policy, by contrast, aggressively prunes ``one-shot'' facts (such as specific timestamps) that lack repetition, resulting in a -44.9\% drop in Temporal recall. Yet, MemArchitect outperformed MemOS in \textbf{Open Domain (+11.3\%)} and \textbf{Multi-Hop (+4.1\%)} tasks. This indicates that while active governance sacrifices eidetic recall, it significantly enhances abstract synthesis and long-horizon coherence. It shifts the failure mode from dangerous ``Context Pollution'' to tunable ``Over-Pruning'', which is a manageable parameter rather than a structural flaw.
\section{Conclusion and Future Work}
\label{sec:conclusion}
We introduced \textbf{MemArchitect}, a middleware layer transforming memory from passive storage to active adjudication. By enforcing policies for lifecycle, consistency, retrieval, and efficiency, we address the critical ``Governance Gap.'' Our prototype demonstrates a strong aggregate improvement (+7.45\%) and outperforms SimpleMem (+27.5\% in subgroup analysis), proving that adjudication preserves semantic integrity better than compression. While aggressive decay currently limits raw recall compared to the maximalist baseline, MemOS, these results validate our thesis: Governance shifts failure modes from unchecked hallucination to managed pruning.

To advance MemArchitect from a prototype into a trustworthy and robust policy-driven architecture, we will continue to evaluate the system on established benchmarks such as \textbf{LongMemEval} \citep{wu2024longmemeval}, which challenges model capacity with significantly more question-answer turns than LoCoMo. In parallel, we aim to validate the prototype on persona-centric benchmarks like \textbf{PreFEval} \citep{zhao2025llmsrecognizepreferencesevaluating} and \textbf{PersonaMem} \citep{jiang2025know} to ensure stable, long-horizon identity preservation. Furthermore, we will fine-tune the architecture by calibrating FSRS for low-frequency fact retention, implementing the remaining governance policies, and expanding our evaluation to include comparisons against other state-of-the-art and learning-based approaches for long-term memory management in agentic environments.
\bibliography{refrences}
\bibliographystyle{plainnat}

\appendix

\section{GOVERNANCE LAYER: THE COMPLETE POLICY LAYER}

\label{app:policies}

This appendix provides the complete technical specifications for the MemArchitect policy engine. We categorize policies into four functional domains: Lifecycle, Consistency, Retrieval, and Safety. Each entry details the mathematical formulation, critical constants derived from our implementation, and the research rationale.

\subsection{Domain A: Lifecycle \& Hygiene (The ``Forgetting'' Engine)}
\textbf{Goal:} To actively manage memory retention based on biological decay and information density.

\paragraph{Policy 1: FSRS Retrievability (Active)}
We implement the Free Spaced Repetition Scheduler (FSRS v4) to model the probability of recall ($R$) over time (\citep{ye2022fsrs}).
\begin{equation}
    R(t) = \left(1 + \frac{19}{9} \cdot \frac{t}{S}\right)^{-1}
\end{equation}
\begin{itemize}
    \item \textbf{Constant The$\frac{19}{9} (\approx 2.11)$:} Empirically calibrated on millions of flashcard reviews. It ensures that when elapsed time ($t$) equals stability ($S$), retrievability drops to exactly $32\%$ ($R \approx 0.32$), which is identified as the optimal forgetting index for efficient relearning.
\end{itemize}

\paragraph{Policy 2: Stability Update (Active)}
Successful retrieval strengthens a memory's stability ($S$), which determines how long it will persist.
\begin{equation}
    S_{new} = S_{old} \times \left(1 + 0.5 \times (11 - D) \times (e^{1.5(1-R)} - 1)\right)
\end{equation}
\begin{itemize}
    \item \textbf{Growth Factor (0.5):} Matches FSRS `w8` parameter for balanced growth, preventing memories from becoming ``immortal'' too quickly.
    \item \textbf{Desirable Difficulty ($e^{1.5(1-R)} - 1$):} The exponent $1.5$ creates a non-linear bonus for ``close calls.'' If a memory is retrieved when $R \approx 0.1$ (almost forgotten), it receives a massive $2.86\times$ stability boost compared to easy recalls.
\end{itemize}

\paragraph{Policy 3: Entropy-Triggered Consolidation (Active)}
A background process monitors the information density of the episodic log to trigger compression cycles.
\begin{equation}
    \text{Trigger} \iff \frac{\text{Size}(\text{Gzip}(\text{Content}))}{\text{Size}(\text{Content})} < 0.4
\end{equation}
\begin{itemize}
    \item \textbf{Threshold (0.4):} Text streams compressing to $<40\%$ of original size indicate high redundancy ($>60\%$ repetition).
    \item \textbf{Action:}
    \begin{itemize}
        \item \textbf{Delete ($R < 0.3$):} Memories with $<30\%$ retrievability are effectively forgotten and pruned.
        \item \textbf{Consolidate ($0.3 \le R \le 0.7$):} Fading memories are synthesized into semantic facts.
        \item \textbf{Keep ($R > 0.7$):} Active memories are preserved.
    \end{itemize}
\end{itemize}

\subsection{Domain B: Consistency \& Truth (The ``Utility'' Engine)}
\textbf{Goal:} To maximize factuality and reduce hallucinations through active monitoring.

\paragraph{Policy 4: Kalman Utility Filter (Active)}
We estimate the ``True Utility'' ($U$) of a memory source by treating usage feedback as a noisy signal.
\begin{equation}
    K_k = \frac{P_{k-1}}{P_{k-1} + R_{noise}}; \quad U_{k} = U_{k-1} + K_k \cdot (z_k - U_{k-1})
\end{equation}
\begin{itemize}
    \item \textbf{Process Noise ($Q=0.05$):} Assumes memory utility changes slowly over time.
    \item \textbf{Measurement Noise ($R=0.1$):} Reflects that our Usage Detection heuristic (word overlap) has moderate uncertainty.
    \item \textbf{Kalman Gain ($K \approx 0.6$):} The system weighs new observations at $60\%$ and historical inertia at $40\%$.
\end{itemize}

\paragraph{Policy 5: Relevance Discriminator (Active)}
A Cross-Encoder model acts as a ``Veto Gate'' after vector retrieval to filter high-similarity but irrelevant noise.
\begin{equation}
    \text{Entailment}(Q, M) < 0.1 \implies \text{Drop}
\end{equation}
\begin{itemize}
    \item \textbf{Threshold (0.1):} Acts as a safety floor to discard candidates that are semantically disjoint, even if they passed the vector search.
\end{itemize}

\paragraph{Policy 6: Conflict Resolution (Planned)}
Future work involves an NLI-based arbitrator. When $M_{new}$ conflicts with $M_{old}$:
\begin{equation}
    \text{Winner} = \max(\text{SourceAuth}(M) \times \text{Recency}(M))
\end{equation}
This policy will resolve contradictions (e.g., ``User lives in NYC'' vs ``User lives in London'') to prevent ``schizophrenic'' agent responses.

\subsection{Domain C: Adaptive Retrieval (The ``Auction'' Engine)}
\textbf{Goal:} To dynamically value information based on user intent and task complexity.

\paragraph{Policy 7: Adaptive Scoring (Active)}
The system assigns a Retrieval Score based on similarity ($S$), retrievability ($R$), and utility ($U$).
\begin{equation}
    \text{Score} = \text{Sim} \times R^{\lambda} \times (1 + \beta \cdot U)
\end{equation}
\begin{itemize}
    \item \textbf{Decay Penalty ($\lambda$):}
    \begin{itemize}
        \item $\lambda=0$ (Fact): Facts are timeless; age is ignored.
        \item $\lambda=1.0$ (Reasoning): Immediate context is prioritized; old memories are penalized.
    \end{itemize}
    \item \textbf{Utility Boost ($\beta$):}
    \begin{itemize}
        \item $\beta=1.5$ (Multi-hop): Trust proven memories ($U$) for complex reasoning chains.
    \end{itemize}
\end{itemize}

\paragraph{Policy 8: Hebbian Graph Expansion (Active)}
We track co-occurrence to solve the ``missing link'' problem. If memory $A$ is retrieved, memory $B$ is also retrieved if:
\begin{equation}
    P(B|A) = \frac{\text{Count}(A \cap B)}{\text{Count}(A)} > 0.7
\end{equation}
\begin{itemize}
    \item \textbf{Threshold (0.7):} Requires a $70\%$ co-occurrence rate to establish a strong associative link, filtering out random noise.
\end{itemize}

\paragraph{Policy 9: Adaptive Token Budgeting (Active)}
Token allocation varies with the confidence of the retrieval set.
\begin{equation}
    \text{GenReserve} = \begin{cases} 
    2048 & \text{if } \text{AvgScore}(\text{Top5}) > 0.4 \quad (\text{Reasoning Mode}) \\
    300 & \text{otherwise} \quad (\text{Recall Mode})
    \end{cases}
\end{equation}
\begin{itemize}
    \item \textbf{Reasoning Mode:} High-quality retrieval allows the LLM to utilize a large context for analysis.
    \item \textbf{Recall Mode:} Low-quality retrieval forces a maximized context window to find needle-in-haystack facts.
\end{itemize}

\subsection{Domain D: Governance \& Safety (The ``Compliance'' Engine)}
\textbf{Goal:} To enforce safety standards and data privacy.

\paragraph{Policy 10: Toxic Memory Filter (Planned)}
A write-path guardrail scans all incoming user text. If ``Prompt Injection'' or toxic content is detected, the system refuses to write the entry to long-term storage, preventing persistent poisoning attacks.

\paragraph{Policy 11: ``Right-to-be-Forgotten'' Cascade (Planned)}
To comply with GDPR, this policy ensures that deleting a root memory triggers a recursive purge of all derived summaries and insights. This prevents the agent from retaining ``Zombie Memories''-secrets that persist in summarized forms after the user has requested deletion.

\section{Evaluation Setup}
\label{app:experiments}

To ensure a fair and direct comparison, we conducted two distinct evaluation runs, strictly matching the backbone models and configurations of each respective baseline.

\subsection{Experiment 1: MemOS Comparison Setup}
For the MemOS comparison, we followed the official MemOS evaluation pipeline and reproduced its experimental setup using the same foundation models and evaluation scripts. Both MemOS and MemArchitect were evaluated under identical backbone and embedding configurations to isolate the effect of memory governance.

\begin{itemize}
    \item \textbf{Backbone LLM:} We used \textbf{Meta-Llama-3.1-8B-Instruct} as the reasoning and generation model for both systems.
    \item \textbf{Embedding Model:} Both systems employed \textbf{BGE-M3} for dense retrieval and semantic similarity.
    \item \textbf{Evaluation Protocol:} We ran the original MemOS evaluation script on the LoCoMo benchmark and then evaluated MemArchitect using the same script, prompts, and judging logic, replacing only the memory backend. Answer correctness was assessed using an LLM-as-a-Judge based on Llama-3.1-8B, following MemOS’s reported methodology.
\end{itemize}
This controlled setup ensures that the performance differences reported in Table \ref{tab:merged_results} arise from the memory architecture and governance policies, rather than differences in models or evaluation procedures.

\subsection{Experiment 2: SimpleMem Comparison Setup}
We evaluate Memory Architect against SimpleMem under a strictly controlled setup to isolate the impact of the memory architecture. Both systems use the same backbone LLM, embedding model, dataset, and evaluation protocol, differing only in their memory management strategy.

\begin{itemize}
    \item \textbf{Models:} Both systems use \textbf{Qwen2.5-3B-Instruct} as the generative backbone and \textbf{nomic-embed-text-v1.5} for dense retrieval. To ensure consistency, the same LLM (Qwen2.5-3B-Instruct) is also used as an LLM-as-a-Judge.
    \item \textbf{Evaluation Protocol:} We follow the official SimpleMem LoCoMo-10 evaluation script and apply the exact same judging logic to Memory Architect, replacing only the memory backend. Answer correctness is determined via an LLM-as-a-Judge using a binary pass/fail decision based on semantic match, core fact preservation, and acceptable temporal or lexical variance.
\end{itemize}

\subsection{Dataset Configuration}
Both experiments utilized the \textbf{LoCoMo-10} dataset, a long-horizon conversational benchmark consisting of 1,986 QA pairs across multi-session dialogues. Results are reported by category (Single-Hop, Temporal, Multi-Hop, and Open-Domain) to highlight specific architectural strengths.

\end{document}